\documentclass[10pt,twocolumn,letterpaper]{article}

\usepackage[pagenumbers]{cvpr} 

\usepackage{amsmath}
\usepackage{algorithm}
\usepackage{algpseudocode}
\usepackage{graphicx}
\usepackage{caption}
\usepackage{subcaption}
\usepackage{adjustbox}
\usepackage{array}
\usepackage{colortbl}
\usepackage{multirow}
\usepackage{cuted}
\usepackage{capt-of}

\newcommand{\point}{\mathbf{C}}

\definecolor{cvprblue}{rgb}{0.21,0.49,0.74}
\usepackage[pagebackref,breaklinks,colorlinks,allcolors=cvprblue]{hyperref}

\def\paperID{16761} 
\def\confName{CVPR}
\def\confYear{2025}

\title{Flash3D: Super-scaling Point Transformers through Joint Hardware-Geometry Locality}

\author{
Liyan Chen$^{1}$, Gregory P. Meyer$^{2}$, Zaiwei Zhang$^{2}$, Eric M. Wolff$^{2}$, Paul Vernaza$^{2}$\\
$^{1}$The University of Texas at Austin\\
$^{2}$Cruise LLC\\
{\tt\small liyanc@cs.utexas.edu, eric.wolff@getcruise.com, paul.vernaza@getcruise.com}\\
{\tt\small \url{https://github.com/liyanc/Flash3DTransformer}}
\vspace{-0.13in}
}

\begin{document}
\maketitle
\begin{abstract}
Recent efforts recognize the power of scale in 3D learning (e.g. PTv3) and attention mechanisms (e.g. FlashAttention).
However, current point cloud backbones fail to holistically unify geometric locality, attention mechanisms, and GPU architectures in one view.
In this paper, we introduce \textbf{Flash3D Transformer}, which aligns geometric locality and GPU tiling through a principled locality mechanism based on Perfect Spatial Hashing (PSH).
The common alignment with GPU tiling naturally fuses our PSH locality mechanism with FlashAttention at negligible extra cost.
This mechanism affords flexible design choices throughout the backbone that result in superior downstream task results.
Flash3D outperforms state-of-the-art PTv3 results on benchmark datasets, delivering a 2.25x speed increase and 2.4x memory efficiency boost. 
This efficiency enables scaling to wider attention scopes and larger models without additional overhead.
Such scaling allows Flash3D to achieve even higher task accuracies than PTv3 under the same compute budget.
\end{abstract}

\vspace{-0.08in}
\section{Introduction}
\label{sec:intro}
Efficient and scalable processing of point clouds is crucial for a wide range of applications, including autonomous driving~\cite{sun2020scalability,hu2022processing,hu2023planning,Ye2023FusionADMF,Ngiam2021SceneTA}, robotic navigation~\cite{LobosTsunekawa2020PointCB,Qin2022DexPointGP,Yan2019DataEfficientLF,Yoo2023TowardZS,Ugurlu2022SimtoRealDR}, and augmented reality~\cite{Qiu2021SemanticSF,Franzluebbers2022VirtualRP}. 
Point cloud backbones are essential architectures that extract meaningful features from raw 3D data. 
Recent advancements~\cite{wu2024point, liu2023flatformer, wang2023octformer, lai2023spherical,lai2022stratified} have explored various strategies to enhance the performance of these backbones, particularly focusing on how Multi-Head Self-Attention (MHSA) mechanisms can be optimized for 3D point clouds.
Windowing MHSA and region shifting (e.g., Swin Transformers~\cite{liu2021swin,yang2023swin3d}) enable efficient local-global feature extraction, but scaling point cloud backbones to larger point cloud sizes and model capacities remains challenging~\cite{liu2021swin,wu2024point,wang2023octformer}.

\begin{figure}[h]
   \vspace{-0.12in}
    \centering
    \begin{subfigure}[t]{\columnwidth}
        \centering
        \includegraphics[width=\textwidth]{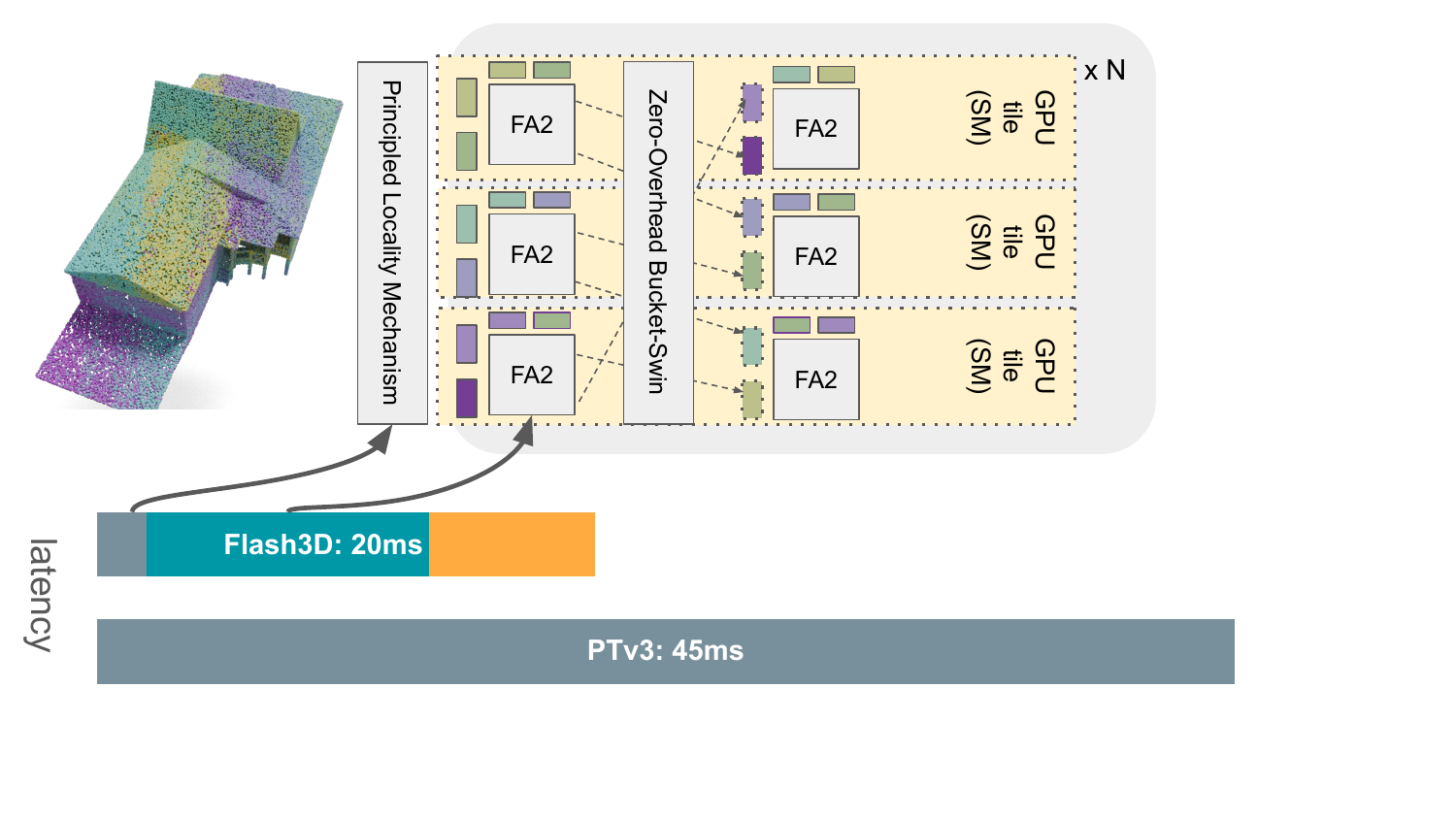} 
        \label{fig:teaser_row1}
    \end{subfigure} 
    
    \vspace{-0.12in}
    \caption{Effectiveness our Flash3D transformer by unifying geometric locality, FlashAttention (FA2), and GPU tiling architecture. Our unified perspective leads to drastically improved speed and scalability of point transformers.}
    \label{fig:teaser}
    \vspace{-0.18in}
\end{figure}

In this paper, we tackle the challenges of efficient and scalable point cloud backbones by introducing \textbf{Flash3D Transformer}, a new architecture that unifies geometric locality with GPU memory locality. This is achieved through a principled locality mechanism based on Perfect Spatial Hashing (PSH). Flash3D is designed to align point cloud backbones with the tiling architecture of GPUs, leading to marked improvements in both efficiency and scalability. 
Our work integrates a locality mechanism that bridges geometric and GPU memory locality, enabling an efficient mapping of 3D points into compact memory spaces. 
Benefiting from compact memory layouts, Flash3D introduces a multilevel attention grouping method, \textbf{Bucket-and-Swin}, precisely aligned with GPU tiling, allowing zero overhead region shifting fused with FlashAttention-2~\cite{dao2023flashattention}.
The joint locality and efficiency boost are illustrated in Figure~\ref{fig:teaser}.

Our key contributions are summarized as follows:

\begin{itemize}
  \item \textbf{Unified geometric and GPU memory locality}: Using Perfect Spatial Hashing (PSH), we propose a principled locality mechanism that brings together geometric and memory locality, benefiting downstream feature quality and operation throughput.
  
  \item \textbf{Attention aligned with GPU tiling}: We propose a novel \textbf{Bucket-and-Swin} attention mechanism, structured closely to GPU tiling, incorporating zero-overhead region shifting and striding.


  \item \textbf{Scalable Performance}: Flash3D Transformer outperforms PTv3 on benchmark datasets, simultaneously achieving a 2.25x speed increase and 2.4x memory efficiency boost.
\end{itemize}
\begin{figure*}[ht]
\centering
\includegraphics[width=0.9\textwidth]{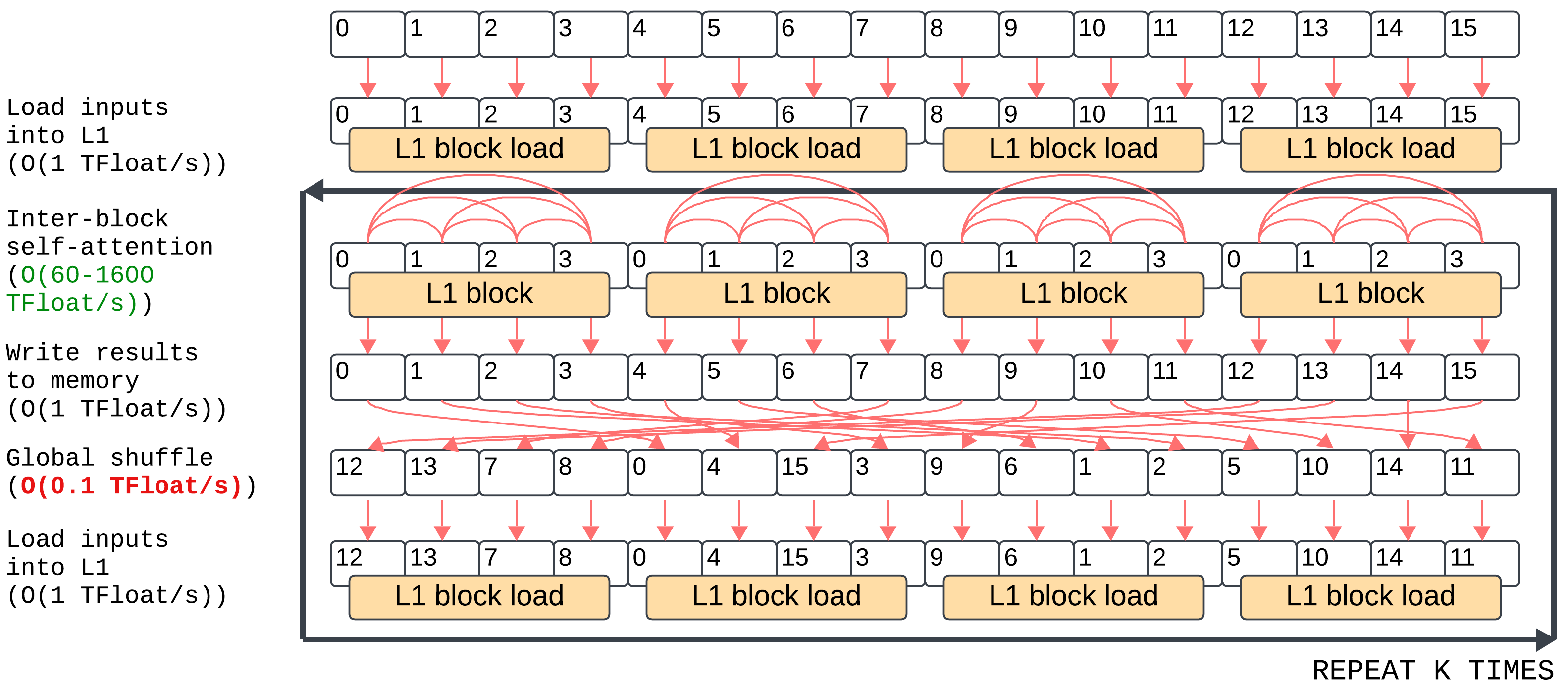}

\vspace{-0.08in}
\caption{High-level schematic overview of PTv3. Numbered rectangles represent locations in memory. Adjacent rectangles are adjacent in memory. Arrows indicate data movement. See Section~\ref{sec:prelim} for details.}
\label{fig:ptv3_overview}

\end{figure*}

\begin{figure*}
\centering
\includegraphics[width=0.9\textwidth]{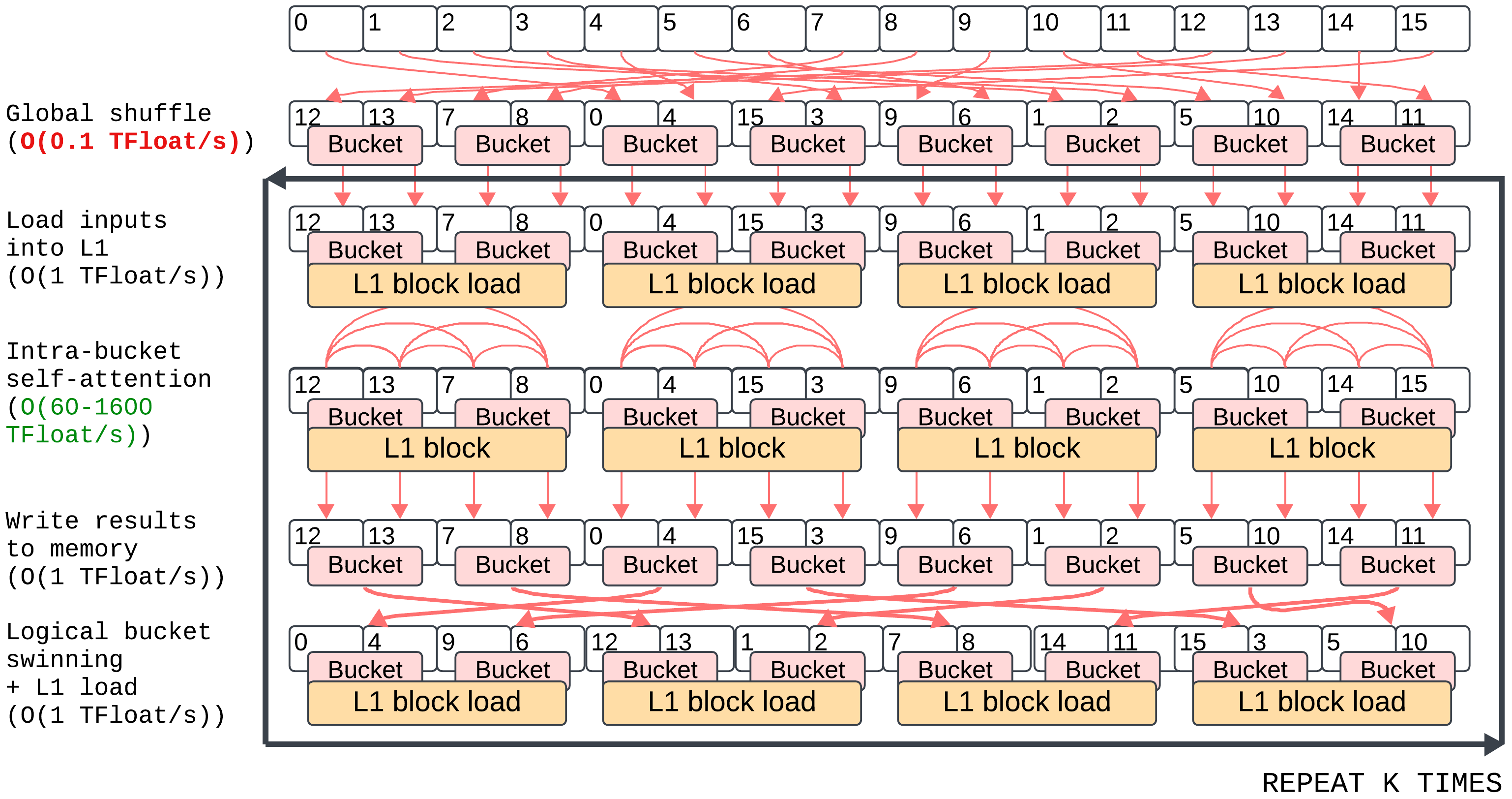}

\vspace{-0.08in}
\caption{High-level schematic overview of Flash3D. Flash3D performs multiple rounds of attention with different neighborhood definitions via our {\em bucket-and-swin} approach, which saves an expensive global shuffle in each round. See Section~\ref{sec:method} for details.}
\label{fig:flash3d_dataflow_overview}

\vspace{-0.15in}
\end{figure*}

\section{Related Work}

\paragraph{3D Point Transformers}
Many variations of Point Transformers have appeared recently, such as PTv1~\cite{zhao2021point}, PTv2~\cite{wu2022point}, PTv3~\cite{wu2024point}, FlatFormer~\cite{liu2023flatformer}, and OctFormer~\cite{wang2023octformer}, each differing in their definitions of geometric locality. 
PTv1 and PTv2 introduced foundational concepts in point cloud transformers, focusing on capturing both local and global geometric structures within point clouds through intricate windowing of MHSA. 

FlatFormer~\cite{liu2023flatformer} demonstrated that partitioning point clouds into equally sized windows benefits MHSA by enabling efficient batching. 
OctFormer~\cite{wang2023octformer} proposed octree-based neighborbood partitioning for fast window computation.
Additionally, region shifting based on Swin Transformers has been utilized to aggregate and distribute global information through subsequent local windowing, enhancing the network's ability to capture both local and global features~\cite{liu2021swin,fan2022embracing,liu2023flatformer}. PTv3~\cite{wu2024point} incorporated such design principles and achieved state-of-the-art performance by scaling up MHSA window sizes and parameter sizes.
However, by relying on abstractions of global serialization and scattering of points and features, PTv3 overlooks GPU architectural issues, incurring substantial and unnecessary computational and memory costs (see Figure~\ref{fig:ptv3_overview} and Section~\ref{sec:prelim} for details).  This highlights the need to holistically integrate geometric locality with hardware considerations.

Varied definitions of geometric locality call for a general approach to encompass a family of such. Ideal geometric locality definitions should be flexible and compute-efficient to represent striding and shifting and incorporate global structures within point clouds.
Variations among partition sizes lead to negligible differences in downstream task performances while hindering the computing efficiency~\cite{wang2023octformer,sun2022swformer}. 
Therefore, we motivate our principled locality mechanism to align flexible geometric locality definitions with evenly sized windows, which allow regularly batched computations on GPUs and saturate GPU computing throughput. 
Our \textbf{Bucket-and-Swin} attention offers an efficient method for striding and shifting on top of properly localized points by leveraging our principled locality mechanism.

\vspace{-0.08in}
\paragraph{FlashAttention}
FlashAttention algorithms~\cite{dao2022flashattention,dao2023flashattention,shah2024flashattention} leverage the tiling structure of GPU chips and optimize attention mechanisms by retaining intermediate results on-chip and carefully localizing computations. 
Retention of intermediate results on-chip helps FlashAttention algorithms to reduce quadratic memory cost down to a linear one~\cite{milakov2018online,shen2021efficient,dao2022flashattention,spector2024thunderkittens}.
Partitioning input arrays into tiles aligned with GPU tiling structures maximizes computation throughput.
Given that GPU architectures have maintained a tiling structure for over two decades—a fundamental design aspect unlikely to change~\cite{10.5555/3207796}—these methods aim to maximize GPU utilization for a range of GPU chips.

These two principles motivate our design choices of \textbf{Bucket-and-Swin} attention.
We partition point cloud features into tiles fit in GPU tiles; each tile is fetched on-the-fly based on logical assignments; logical partitions and fetching are fused with FlashAttention-2 CUDA kernel.

\vspace{-0.08in}
\paragraph{Perfect Spatial Hashing (PSH)}
Spatial hashing was originally proposed to efficiently locate and query spatially sparse data points~\cite{lo1996spatial,fredman1984storing}. To enhance memory locality and computational efficiency, researchers introduced the concept of \emph{perfectness}~\cite{czech1997perfect}, compacting the hash table into a contiguous linear array in memory. This approach resulted in improved memory access patterns and query speed~\cite{lefebvre2006perfect}. Later, \cite{alcantara2009real} proposed a parallelized PSH construction on GPUs, achieving exceptionally fast constructions even on early GPUs by leveraging their massive parallel processing capabilities and placing buckets within GPU tiles.

In our work, we focus on perfectness, propose a principled locality mechanism, and remove the traditional table indexing structure. 
We focus on hashing a large number of points into a contiguous memory array for MHSA. 
This simplification aligns with our goal of unifying geometric locality with GPU memory locality to optimize performance.
\section{Preliminaries}
\label{sec:prelim}

In order to explain how Flash3D achieves fast performance, we must first understand the bottlenecks in previous work. Figure~\ref{fig:ptv3_overview} depicts the high-level dataflow associated with a method such as PTv3 (note that some layers are omitted for the sake of simplicity of exposition). Each numbered rectangle represents a float stored in memory. We assume the data has been pre-sorted to ensure contiguous floats are likely to be close in terms of some meaningful metric.

Before the data can be processed, it must be loaded from main memory into the L1 cache (which is practically equivalent to shared memory in GPU jargon). An H100 can perform this step at a rate of roughly 1 TFloat/s. Although this seems like an impressive number, it is dwarfed by the maximum achievable FLOP throughput of the GPU, which is in excess of 60 TFloat/s. In our case, the compute-intensive step consists of FlashAttention, which can perform in excess of 500 TFLOPs/s (albeit at reduced precision).

After writing results to memory, PTv3 then performs a global reordering of the points in order to communicate information across a different set of neighborhoods. As depicted in the figure, this step is orders of magnitude slower than the other steps. This is because the GPU is optimized to transfer large amounts of data in transactions of contiguous blocks. Transferring data between random locations is analogous to using many buckets (memory transactions) to move water one drop (float) at a time---this process can only be efficient if we fill the buckets (by transferring the data in contiguous chunks)\footnote{Though global shuffling through raw DRAM channels bottlenecks at 0.1TFloat/s, Flash3D effectively coalesces shuffling in L2 cache and approximates 1TFloat/s limit in reality. One million points $\texttt{FP16}[1\text{M}, 3]$ cost $6\text{MB}$ memory while H100 has $50\text{MB}$ L2 cache to coalesce shuffled results, write well-packed results, and maximize DRAM bandwidth.}.

\vspace{-0.12in}
\section{Method}
\label{sec:method}

From the preceding discussion, it is clear that the global shuffle step of PTv3 is a key bottleneck. Our key idea is to mitigate this bottleneck through our \textbf{Bucket-and-Swin} strategy. In this approach, illustrated in Figure~\ref{fig:flash3d_dataflow_overview}, we initially bucket the points into spatially similar neighborhoods using hash functions. This involves an up-front global shuffle operation that places points in the same contiguous block of memory if they are in the same bucket.

In subsequent iterations, instead of repeating a slow global shuffle in order to communicate information using different neighborhoods, we logically shuffle the buckets to yield different neighborhoods for attention. This shuffle is logical in the sense that it does not involve any permutation of bytes in memory---instead, we simply load the appropriate buckets into L1 just before calculating attentions. To return to the water analogy, this is akin to transferring water using full buckets of water (by loading entire point-buckets into L1) instead of nearly empty buckets. We are thus able to mostly mitigate the global shuffling bottleneck.

Implementing this strategy efficiently involves a few non-trivial details, which we elaborate upon in the following subsections.

\subsection{Principled Locality Mechanism with PSH}

\begin{figure*}[ht]

    \vspace{-0.12in}
    
    \begin{subfigure}[t]{0.24\textwidth}
        \centering
        \adjustbox{height=0.175\textheight}{\includegraphics{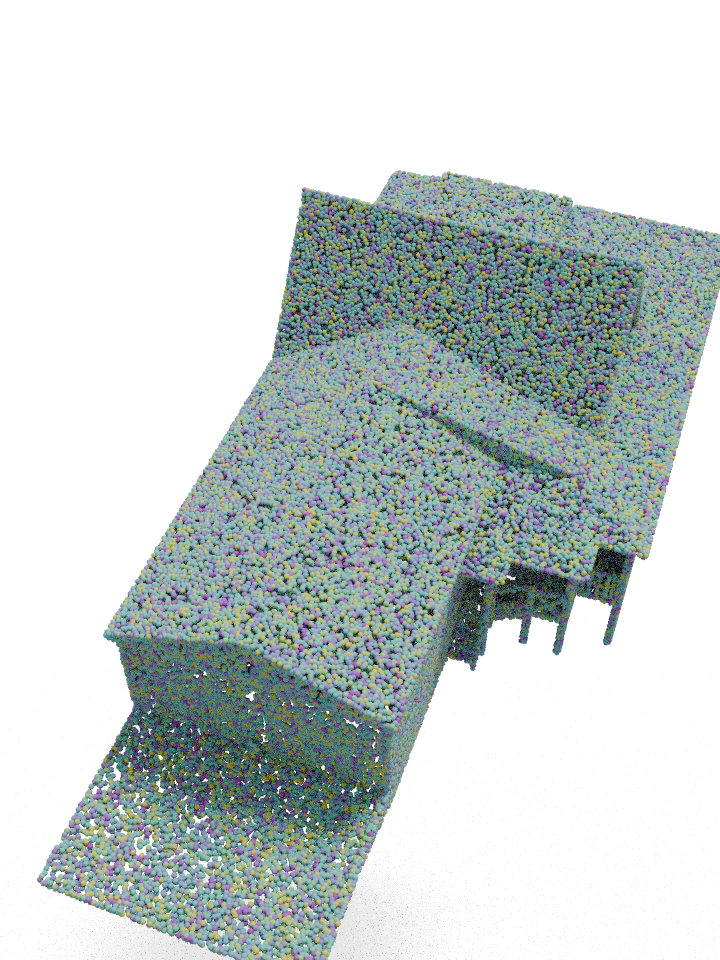}} 
        \caption{XOR-mod (Rebalanced)}
    \end{subfigure}
    \hfill
    \begin{subfigure}[t]{0.24\textwidth}
        \centering
        \adjustbox{height=0.175\textheight}{\includegraphics{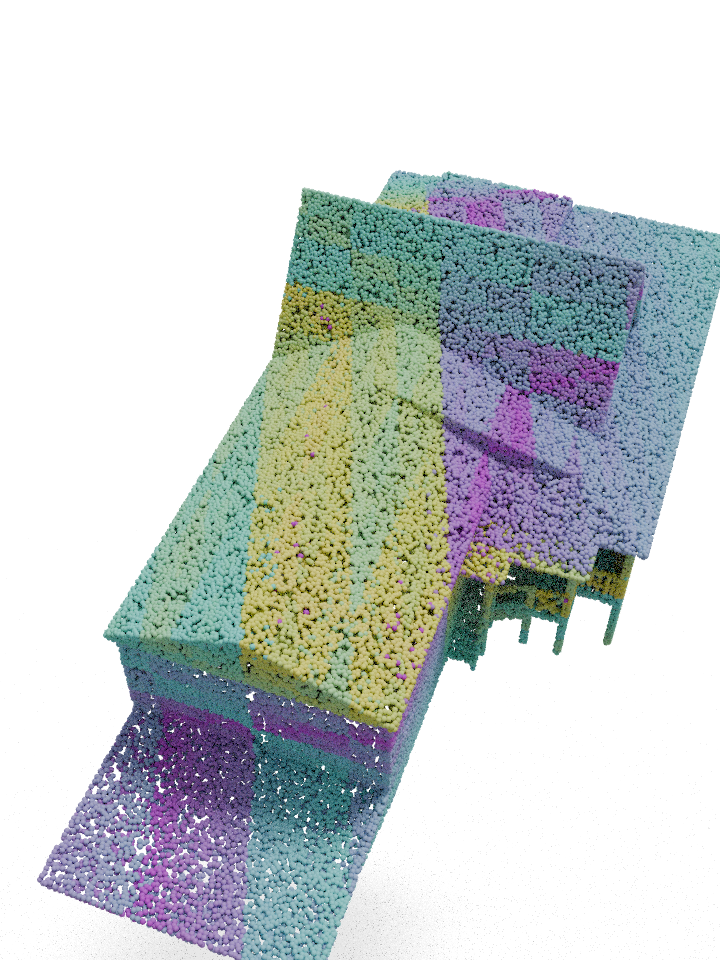}} 
        \caption{XOR-div (Rebalanced)}
    \end{subfigure}
    \hfill
    \begin{subfigure}[t]{0.24\textwidth}
        \centering
        \adjustbox{height=0.175\textheight}{\includegraphics{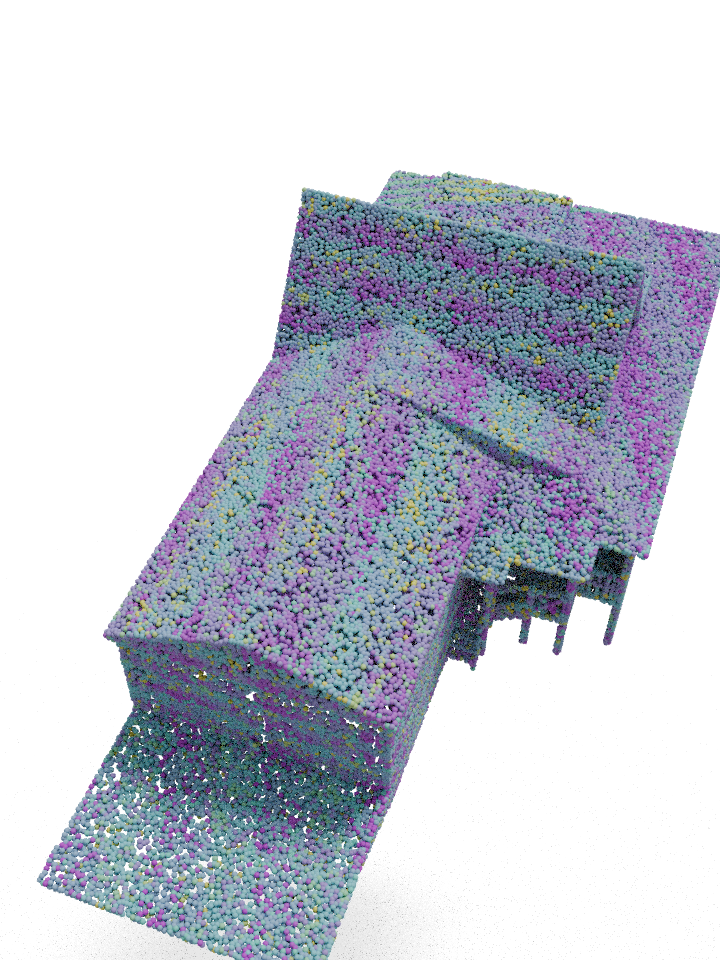}} 
        \caption{Zorder-mod (Rebalanced)}
    \end{subfigure}
    \hfill
    \begin{subfigure}[t]{0.24\textwidth}
        \centering
        \adjustbox{height=0.175\textheight}{\includegraphics{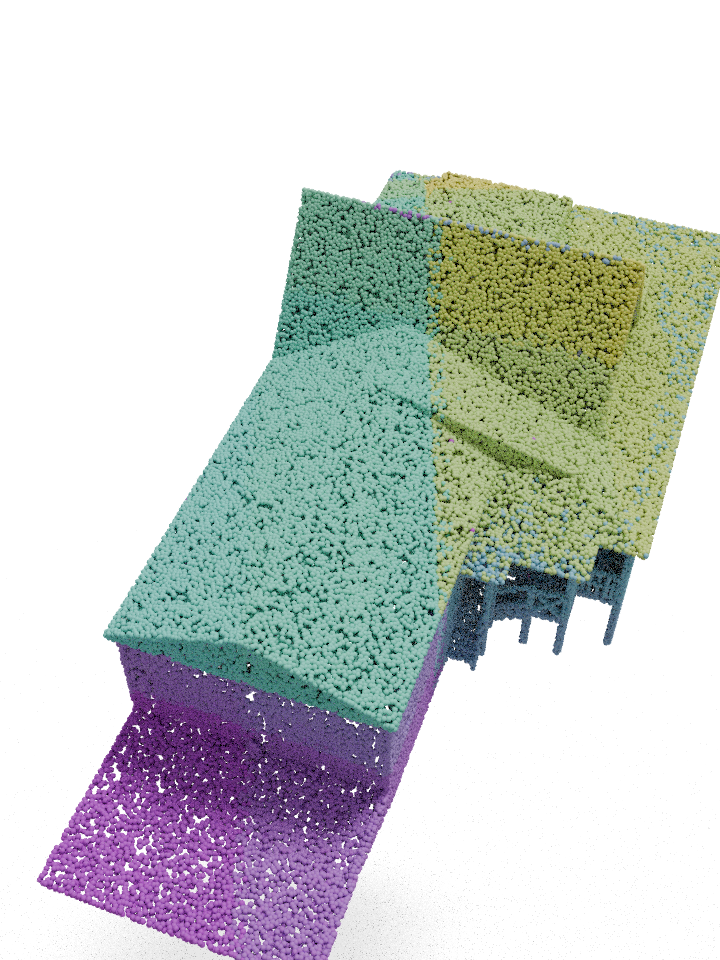}} 
        \caption{Zorder-div (Rebalanced)}
    \end{subfigure}%

    \vspace{-0.08in}
    \caption{Illustration of bucket assignments using four hash functions after rebalancing. Colors of points indicate their bucket assignments. We demonstrate our PSH algorithm on a sample point cloud with 100k points from BuildingNet~\cite{DBLP:conf/iccv/SelvarajuNLMAAC21}.}
    \label{fig:hash_function_buckets}

    \vspace{-0.2in}
    
\end{figure*}
Our principled locality mechanism leverages Perfect Spatial Hashing (PSH)~\cite{lefebvre2006perfect,alcantara2009real} to map spatially sparse 3D point clouds into compact and contiguous arrays. 
PSH constructs a bijection to scatter 3D points. 
In the output array, memory address proximity implies spatial proximity.

\noindent \textbf{Input and Output} The PSH algorithm takes as input the point coordinates \(\mathbf{C} \in \mathbb{R}^{N \times 3}\), representing \(N\) points in 3D space, and the bucket capacity \(S\), defining the maximum points per bucket. 
It outputs the bucket IDs \(\mathbf{bucket\_id} \in \mathbb{Z}^N\), assigning each point to a bucket, and bucket offsets \(\mathbf{bucket\_offset} \in \mathbb{Z}^N\), indexing each point’s position within its assigned bucket. 
The index of the \(i\)-th point can be determined by \(\mathbf{bucket\_base}[\mathbf{bucket\_id}[i]] + \mathbf{bucket\_offset}[i]\). 
When buckets are concatenated back-to-back, the resulting indices are guaranteed to be contiguous, and scattering points according to these indices produces a contiguous array of points. We describe our full PSH construction algorithm in Alg~\ref{alg:psh_main} and Alg~\ref{alg:optimistic_racing}.

\begin{algorithm}[h]
\caption{Batch PSH Bucketing and Balancing}
\label{alg:psh_main}
\begin{algorithmic}[1]
\Require Coordinates $\mathbf{C} \in \mathbb{R}^{N \times D}$, batch indices $\mathbf{B}$, number of buckets $K$, bucket capacity $S$
\Ensure Bucket IDs $\mathbf{bucket\_id}$, Bucket Offsets $\mathbf{bkt\_off}$
\For{each batch $b$ \textbf{in parallel}}
    \State Initialize bucket counters $\mathbf{bkt\_ctr} \leftarrow 0$
    \For{each point $i$ in batch $b$ \textbf{in parallel}}
        \State Compute voxel coordinate $\mathbf{v}_i$ from $\mathbf{C}_i$
        \State Compute initial bucket ID $h_i$ from $\mathbf{v}_i$ using hash function
        \If{$\mathbf{bkt\_ctr}[h_i] < S$}
            \Comment{Assign point $i$ to bucket $h_i$}
            \State $\mathbf{bucket\_id} \leftarrow h_i$
            \State $\mathbf{bkt\_off}[i] \leftarrow$ \Call{AtomicInc}{$\mathbf{bkt\_ctr}[h_i]$}
        \Else
            \State \Call{OptimisticRacing}{$i$, $\mathbf{v}_i$, $\mathbf{bkt\_ctr}$, $S$}
        \EndIf
    \EndFor
    \State Perform exclusive scan on $\mathbf{bkt\_ctr}$ to compute $\mathbf{bucket\_base}$
    \State Scatter points into contiguous memory based on bucket assignments
\EndFor

\end{algorithmic}
\end{algorithm}

\noindent \textbf{Stage Orchestration} In Flash3D, PSH orchestrates three key components within each transformer stage. 
First, PSH groups spatially adjacent points into contiguous memory arrays and produces \(\mathbf{bucket\_id}\), \(\mathbf{bucket\_offset}\). The resulted array suits point-wise MLPs and LayerNorm.
The second component maps buckets to attention scopes during FlashAttention without extra cost.
Lastly, the bucket structure directs a tile-level in-bucket pooling layer. 
Because points within each bucket are geometrically local, pooling operations are confined within buckets, reducing the need for costly global neighbor queries. 
This localized pooling achieves high computational efficiency by retaining geometric coherence in point clusters, enabling efficient, spatially-aware feature aggregation throughout the model.

\noindent \textbf{Bucket rebalancing}
Ideally, we would assign a bucket to each point by simply hashing it. Unfortunately, this would lead to imbalanced buckets, with some buckets containing many points and others containing few points. Such non-uniform buckets would prevent us from efficiently mapping buckets onto GPU tiles.
To avoid this issue, we aim for \textit{best-effort} geometric locality instead of rigidly defining geometric boundaries, allowing the GPU to settle point-to-bucket assignments opportunistically. This declarative approach yields additional advantages.
\begin{algorithm}[h]
\caption{Optimistic Racing for Bucket Reassignment}
\label{alg:optimistic_racing}
\begin{algorithmic}[1]
\Require Point index $i$, voxel coordinate $\mathbf{v}_i$, bucket counters $\mathbf{bkt\_ctr}$, bucket capacity $S$
\Ensure Updated bucket assignment for point $i$
\State Initialize assigned bucket ID $h_{\text{assigned}} \leftarrow -1$
\For{each probe offset $\delta$}
    \State Perturb voxel coordinate $\mathbf{v}'_i \leftarrow \mathbf{v}_i + \delta$
    \State Compute new bucket ID $h'_i$ from $\mathbf{v}'_i$ using hash function
    \If{$\mathbf{bkt\_ctr}[h'_i] < S$}
        \State $\mathbf{prev\_off}$ = \Call{AtomicInc}  {$\mathbf{bkt\_ctr}[h'_i]$}
        \If{$\mathbf{prev\_off} \leq S$} 
        \Comment{Confirm bucket capacity after increment}
            \State Assign point $i$ to bucket $h'_i$
            \State $\mathbf{bucket\_offset}[i] \leftarrow \mathbf{prev\_off}$
            \State $h_{\text{assigned}} \leftarrow h'_i$
            \State \textbf{break}
        \Else
            \State \Call{AtomicDec}{$\mathbf{bkt\_ctr}[h'_i]$}
            \Comment{Roll back if over capacity}
        \EndIf
    \EndIf
\EndFor
\If{$h_{\text{assigned}} = -1$}
    \State Assign point $i$ to recycle bucket $r$
    \State $\mathbf{bucket\_offset}[i] \leftarrow \Call{AtomicInc}{\mathbf{bkt\_ctr}[r]}$
\EndIf

\end{algorithmic}
\end{algorithm}

First, relaxed geometric boundaries allow point-bucket associations to be resolved in on-chip arbitration, significantly boosting grouping throughput. Second, softened boundaries result in diffused geometric patterns, facilitating more flexible feature extraction through attention mechanisms. Randomly perturbed boundaries regularizes attention mechanisms without costly feature shuffling as seen in PTv3~\cite{wu2024point}. We show the effects in Table~\ref{table:hash_swin_variants}.

\noindent \textbf{Hash functions}
Our principled locality mechanism in Flash3D leverages the flexibility of hash functions to define geometric locality. 
This approach benefits from a declarative structure, allowing geometric locality definitions to be managed directly through hash functions. 
In contrast to learning from data, our hash functions are manually defined to capture common geometric patterns effectively, as demonstrated by our empirical results.

In this work, we utilize four hash functions that operate over point coordinates to distribute spatially close points into buckets. 
These include \textbf{XOR-mod}, \textbf{XOR-div}, \textbf{Zorder-mod}, and \textbf{Zorder-div}:

\noindent\textbf{XOR-mod}: This function computes the XOR of the voxel coordinate bits and applies a modulo operation. Given voxel coordinate $\mathbf{v}\in \mathbb{Z}^3$, the XOR-mod hash is defined as:
\vspace{-0.08in}
\[
h(\mathbf{v}) = \left( \bigoplus_{d=1}^3 v_d \right) \mod K
\]
where \( v_d \) represents the $d$-th dimension of $\mathbf{v}$, and $K$ is the number of buckets. XOR-mod evenly spreads the geometric overage of each bucket.

\noindent\textbf{XOR-div}: Similar to XOR-mod, this function computes the XOR of the voxel coordinate bits but uses division to determine the bucket. The XOR-div hash is defined as:
\vspace{-0.08in}
\[
h(\mathbf{v}) = \left( \bigoplus_{d=1}^3 v_d \right) \div S
\]
where \( S \) is a scaling factor that controls the bucket size. XOR-div ensures uniform bucket distribution across large voxel ranges, while keeping tight intra-bucket coherence.

\noindent\textbf{Zorder-mod}: This function calculates a Z-order (Morton)~\cite{morton1966computer} code by interleaving the bits of each dimension in $\mathbf{v}$, then applies a modulo operation:
\vspace{-0.08in}
\[
h(\mathbf{v}) = Z(\mathbf{v}) \mod K
\]
where \( Z(\mathbf{v}) \) represents the interleaved Z-order code of $\mathbf{v}$. Zorder-mod preserves spatial proximity, mapping nearby points to similar buckets.

\noindent\textbf{Zorder-div}: This function also uses Z-order coding but divides the Z-order value to allocate points to buckets:
\vspace{-0.08in}
\[
h(\mathbf{v}) = Z(\mathbf{v}) \div S
\]
where \( S \) is a divisor that adjusts the bucket density. Zorder-div is effective for capturing structured spatial locality over extensive regions.

These hash functions provide the necessary flexibility to define geometric locality without the complexity of learning from data, making them both efficient and versatile. 
Additional hash functions can be defined as needed, allowing Flash3D to adapt to diverse geometric patterns and data distributions.

\subsection{Flash3D Transformer Stage}
\label{sec:method_tsfmr_stage}
Windowed attention mechanisms are well-suited to point clouds, as they capture both local and global features by focusing on spatially close points and shifting windows across regions~\cite{fan2022embracing,liu2023flatformer,wu2022point,wu2024point}. We describe how Flash3D achieves local and global feature extraction with minimal computational costs.

\noindent \textbf{Zero-Overhead Bucket-Swin Attention}
Flash3D introduces a zero-overhead bucket-based Swin attention mechanism to efficiently manage local and global feature extraction within point clouds. By organizing points into geometrically localized buckets, Flash3D enables windowed attention over these buckets, simulating the effect of Swin attention shifts without additional computational cost. This approach achieves both high memory locality and effective feature propagation across local neighborhoods.
Given a collection of feature representations $\{\mathcal{F}_{\mathcal{A}_i}\}$ for an attention scope $i$, the bucket-Swin attention mechanism can be described as follows:
\begin{align}
\mathcal{F}_{\mathcal{A}_i}' &= \text{MHSA}(\text{LN}(\{\mathcal{F}_{\mathcal{A}_i}\}), \text{PE}(\point_{\mathcal{A}_i})) + \mathcal{F}_{\mathcal{A}_i} \notag\\
\tilde{\mathcal{F}}_{\mathcal{A}_i} &= \text{MLP}(\text{LN}(\mathcal{F}_{\mathcal{A}_i}')) + \mathcal{F}_{\mathcal{A}_i}' \notag
\end{align}
\noindent
where $\{\mathcal{F}_{\mathcal{A}_i}\}$ represents the input feature collection over multiple buckets in the attention scope, $\text{MHSA}$ denotes Multi-Head Self-Attention, $\text{LN}$ represents Layer Normalization, and $\text{PE}(\point_{\mathcal{A}_i})$ is the positional encoding based on spatial coordinates, not bucket indices.

To illustrate the zero-cost bucket-Swin shift, consider a list example of bucket indices in a point cloud:

\vspace{-0.13in}
\[
\begin{bmatrix}
1 & 2 & 3 & 4 & 5 & 6 & 7 & 8 
\end{bmatrix}
\]
\noindent
Each attention scope in the initial configuration consists of multiple buckets. 
For instance, an attention scope may include buckets $\{1, 2, 3, 4\}$, allowing points within these buckets to interact. 
By shifting the attention window in a subsequent stage, such as by two buckets, we enable cross-scope interactions. 
For example, a shifted scope for two buckets includes buckets $\{3, 4, 5, 6\}$, thus allowing bucket 4 to exchange information with buckets outside its initial neighborhood.

This bucket-level organization enables Swin-style window shifting without needing to recompute neighborhoods globally. 
Instead, the attention shifts are confined within predefined buckets that preserve geometric locality. 
This approach eliminates the computational overhead associated with traditional Swin attention shifts, achieving ``zero-overhead'' locality shifts that are both memory- and computation-efficient.

The zero-overhead bucket-Swin attention mechanism in Flash3D leverages the inherent geometric structure of point clouds to streamline and accelerate attention operations, ensuring high efficiency and effective feature extraction across point cloud regions.

\noindent \textbf{In-bucket Pooling}
Popular pooling operations like grid pooling in PTv2~\cite{wu2022point} incur significant inter-file communications on GPUs.
However, we can capitalize on the existing geometric locality by our principled locality mechanism and localize pooling operations on GPU tiles with fast in-tile resources.
After scattering points and their features into a contiguous array, any subarray aligned with the bucket size contains points that are geometrically close. 
This alignment allows for a fast pooling approach, capitalizing on the geometrical locality of points within each subarray. 
Given that each bucket-size aligned subarray fits within a GPU tile and its fast local L1 cache (equivalently, shared memory per Section~\ref{sec:prelim}), we can execute a similar bucketing and balancing operation within the tile at higher efficiency than our main PSH algorithm.

Our in-bucket pooling layer uses a fixed reduction factor $\rho$, which corresponds to the capacity of each sub-bucket. 
For instance, setting $\rho = 2$ enables 2x pooling, where two points are reduced into one. Typically, we load 1024 points into a GPU tile, construct sub-buckets within the tile, and then reduce the corresponding features using a reduction operation (e.g., \texttt{sum}, \texttt{mean}, \texttt{min}, \texttt{max}, etc.). 
Finally, we perform a bulk memory transaction to write out the reduced features, minimizing memory access overhead.

In contrast to PTv3, which pivots their pooling operations on global sorting of the whole point cloud, our in-bucket pooling further capitalizes on the geometric locality and memory locality to eliminate the latency costs.

For a detailed explanation of the in-bucket sub-bucket construction and balancing, please refer to the Appendix.

\begin{table}[h!]
    \centering
    \small 
    \setlength{\tabcolsep}{6pt} 
    \arrayrulecolor[gray]{0.2} 
    \begin{adjustbox}{max width=\columnwidth} 
        \begin{tabular}{>{\columncolor[gray]{0.92}}l c c c c}
            \specialrule{1.5pt}{0pt}{0pt} 
            Scalability & \multicolumn{4}{c}{A100} \\
            (nuScenes) & Params. & Memory & Latency & mIoU  \\
            \hline
            PTv3~\cite{wu2024point}    & \cellcolor[gray]{0.8} 46.2M & 1.2G & 45ms          & 80.4         \\
            Flash3D & \cellcolor[gray]{0.8} 46.2M & 0.5G & \textbf{20ms} & \textbf{81.2}\\
            \hline
            PTv3~\cite{wu2024point}    & 46.2M     & \cellcolor[gray]{0.8} 1.2G & 45ms & 80.4          \\
            Flash3D & 129.4M    & \cellcolor[gray]{0.8} 1.2G & \textbf{24ms} & \textbf{81.5} \\
            \specialrule{1.5pt}{0pt}{0pt} 
        \end{tabular}
    \end{adjustbox}
    \caption{Model scalability comparisons of PTv3 and Flash3D on A100 GPUs by fixing model parameter sizes and memory quotas respectively. Dark cells indicate fixed budgets. We fix attention scopes of all models at 4096. mIoU indicates the semantic segmentation performance on nuScenes validation set.}
    \label{table:gpu_comparison}

    \vspace{-0.16in}
\end{table}

\vspace{-0.08in}
\section{Experiments}
\label{sec:exp}
We present empirical evaluations on both \textit{downstream task performance} and \textit{scalability}. With task performance and ablations, we show the impact of various design choices stated in Section~\ref{sec:method}. In the second subsection, we validate hypotheses from Section~\ref{sec:prelim} and further detail the strong scalability resulted from our hypotheses. We report evaluation metrics on a single A100 GPU for fair and generalizable comparisons. 
For results on H100 GPUs see Appendix.
We used \textsc{ThunderKittens}~\cite{spector2024thunderkittens} to implement our fused \textbf{Bucket-and-Swin} attention. For more implementation details see Appendix.

\subsection{Outdoor 3D Task}
We benchmark Flash3D on the nuScenes semantic segmentation task~\cite{caesar2020nuscenes,fong2022panoptic} against a previous state-of-the-art method in Tab~\ref{table:gpu_comparison}.
We train two variants of Flash3D under two sets of quotas to show its performance scalability. 
In the first case, we keep the number of parameters same across Flash3D and PTv3~\cite{wu2024point} shown in the upper half of Tab~\ref{table:gpu_comparison}.
Flash3D outperforms PTv3 on mIoU for 1.0\% at 2.25x inference speed. 
PTv3 uses 2.4x amount of memory than Flash3D.
In the second case, we fix the memory quota for Flash3D and PTv3 shown in the lower half of Tab~\ref{table:gpu_comparison}.
With the same memory budget, Flash3D accommodates 2.8x model parameters and outperforms PTv3 on mIoU by 1.4\%.
Notably, even with 2.8x model parameters, Flash3D inference latency is still 1.88x faster than PTv3.
These results indicate the robust performance scalability of Flash3D. For more downstream task benchmarks see Appendix.

In the following part, we provide ablation analysis of Flash3D on nuScenes validation set.

\begin{table}[h!]
    \centering
    \small 
    \begin{adjustbox}{max width=\columnwidth} 

        \begin{tabular}{>{\columncolor[gray]{0.92}}l c c c c}
            \hline
            Hashs & standard & +rebalance & +stride & +swin \\
            \hline
            XD          & 78.6  & 78.7  & 78.9  & \cellcolor[gray]{0.93} 79.2  \\
            XD+ZD       & 79.1  & 79.1  & 79.8  & \cellcolor[gray]{0.93} 80.6  \\
            XD+XM       & 78.9  & 78.9  & 78.9  & \cellcolor[gray]{0.93} 79.4  \\
            XD+XM+ZD+ZM & 79.3  & 79.4  & 80.2  & \cellcolor[gray]{0.93} \textbf{81.2}  \\
            \hline
        \end{tabular}
    
    \end{adjustbox}

    \vspace{-0.1in}
    \caption{Hash Function Variants and Bucket-based strides and swin on the validation split of the nuScenes dataset. \textsc{+rebalance} adds bucket rebalancing and stochastic geometric boundary perturbations per Alg~\ref{alg:optimistic_racing}. \textsc{+stride} strides attention scopes at two buckets. \textsc{+swin} shifts attention scopes at one bucket a time. }
    \label{table:hash_swin_variants}

    \vspace{-0.12in}
\end{table}

\noindent \textbf{Hash Functions and Rebalancing}
We explore hash and stacked multi-hash scattering strategies and demonstrate their impacts on semantic segmentation tasks in Tab~\ref{table:hash_swin_variants}.
Adding a hash function indicates that we stack a transformer stage led by another hash function. Otherwise, we stack a transformer stage led by the same hash function to keep constant number of stages.
\textbf{XOR-div} gives a reasonable baseline performance and benefits from combining more hash functions in subsequent layers.
Combining \textbf{XOR-div} with a structural hash like \textbf{Zorder-div} gives more performance boost.
Combining \textbf{XOR-div} with a evenly distributed hash like \textbf{XOR-mod} seems give only minor boosts.
When we combine all hash functions defined in Section\ref{sec:method}, Flash3D reaches the highest performance.
We also show the variants of our PSH with and without the rebalancing step per Alg~\ref{alg:optimistic_racing}.
The second column of Tab~\ref{table:hash_swin_variants} indicates that rebalancing seems has little effects on the task performance.

\noindent \textbf{Stride and Swin}
Long-range bucket interactions rely on striding and shifting(swin) patterns. 
We ablate shifting and striding patterns to assess their effectiveness when combining with our stacked multi-hash scattering as in Tab~\ref{table:hash_swin_variants}.
When adding bucket strides, Flash3D gains larger receptive fields and a minor performance improvement.
Adding bucket shifting gives smoother attention scope transitions and further boosts performance.

\subsection{Scalability Analysis}
\begin{figure}[h]
    \centering
    \vspace{-0.16in}
    \includegraphics[width=\linewidth,height=2.5in]{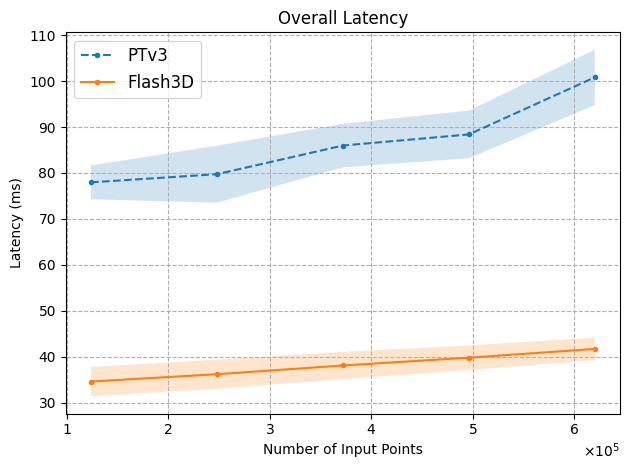}
    
    \vspace{-0.14in}
    \caption{Overall Latencies vs. Input Sizes for Flash3D and PTv3.}
    \label{fig:overall_latency}
    
    \vspace{-0.16in}
    
\end{figure}
As hypothesized in Section~\ref{sec:intro}, scalable point transformers should respect \textit{GPU tiling} and \textit{memory locality}.
In this subsection, we provide detailed profiling results on a single A100 GPU to analyze Flash3D scalability based on these principles.
Specifically, we benchmark Flash3D and PTv3 in terms of latency, compute utilization, and DRAM bandwidth utilization under various input sizes.
For this subsection, we fix Flash3D and PTv3 at the same parameter size.

We vary input sizes from 100k points to 600k points to stress test Flash3D and PTv3. We run both backbones for 20 iterations for each configuration and input size and discard the first warm-up run. With NVIDIA Nsight Systems~\cite{nvidia2024nsight}, we collect tracing logs to report the \texttt{averages} and \texttt{standard deviations} for each metric. 

\vspace{-0.14in}
\subsubsection{Latency}
We measure the overall latencies and latency breakdowns in this section.
We also show the significant latency difference between Flash3D PSH-based principled locality mechanism and serialization algorithm of PTv3.

\begin{figure}[h]
    \centering
    \includegraphics[width=\linewidth]{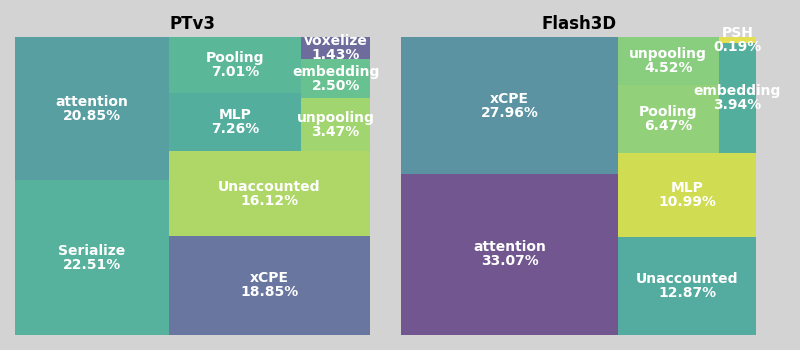}

    \vspace{-0.12in}
    \caption{Latency TreeMap breakdowns for Flash3D and PTv3.}
    \label{fig:treemap}

    \vspace{-0.12in}
\end{figure}
\noindent \textbf{Overall Latency}
We increase the input sizes and show the overall latency difference of Flash3D and PTv3 in Figure~\ref{fig:overall_latency}.
Flash3D consistently out-speeds PTv3 by more than 2x while keeping slower growth and narrower standard deviations.
Flash3D carefully spreads workloads to GPU tiles without global serialization, so Flash3D enjoys a slow linear growth.
On the other hand, global serialization and sorting algorithms in PTv3 incur super-linear complexity and exhibit super-linear latency costs.

\begin{figure}[h]
    \centering
    \vspace{-0.16in}
    \includegraphics[width=\linewidth,height=2.3in]{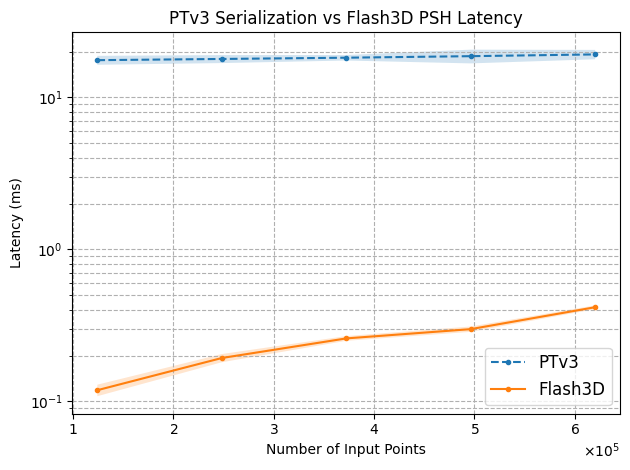}
    
    \vspace{-0.12in}
    \caption{PTv3 Serialization vs. Flash3D PSH, log scale.}
    \label{fig:psh_latency}
    
    \vspace{-0.18in}
\end{figure}
\noindent \textbf{Latency Breakdown with Fixed Input Size}
To further understand the sources of latency, we perform a detailed breakdown of the time spent in different components of Flash3D and PTv3, shown in Figure~\ref{fig:treemap}. 
The global serialization costs a major portion of PTv3 latency while Flash3D PSH latency cost is negligible (0.19\%).

\noindent \textbf{Serialization vs PSH Latency}
We isolate the latencies of serialization and Flash3D PSH in Figure~\ref{fig:psh_latency}.
Flash3D PSH latency is consistently two orders of magnitude lower than PTv3 serialization. 
This difference is a powerful proof of scalability impacts by respecting \textit{GPU tiling}.

\vspace{-0.14in}
\subsubsection{Hardware Utilization Analysis}
We provide an in-depth analysis of hardware utilization to evaluate how efficiently Flash3D utilizes GPU resources during execution. We focus on three key metrics: compute utilization, matrix multiplication utilization, and memory bandwidth utilization.

\begin{figure}[h]
    \centering

    \vspace{-0.14in}
    \includegraphics[width=\linewidth,height=2.1in]{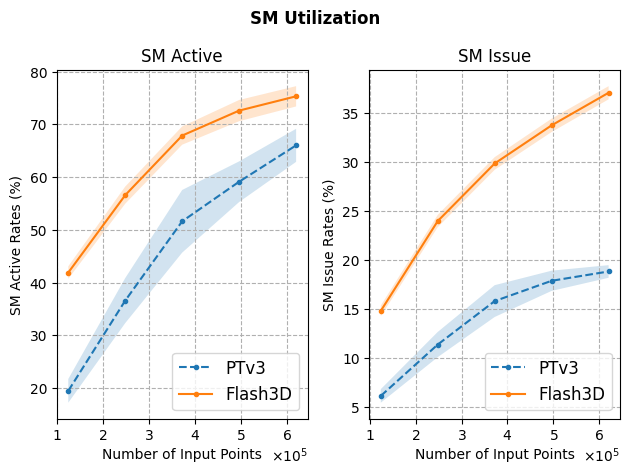}

    \vspace{-0.14in}
    \caption{SM Utilization vs. Input Sizes for Flash3D and PTv3. We show the overall SM active rates on the left and more specific SM issuing rates on the right.}
    \label{fig:sm_active_rates}
    
    \vspace{-0.2in}
\end{figure}
\noindent \textbf{General Compute Utilization}
We report the SM (Streaming Multiprocessor) active rates and SM issuing rates in Figure~\ref{fig:sm_active_rates}. Both measure GPU tile utilization while SM issuing rates are more specific to warp dispatcher instruction throughput.

\begin{figure}[h]
    \centering
    \vspace{-0.16in}
    \includegraphics[width=\linewidth,height=2in]{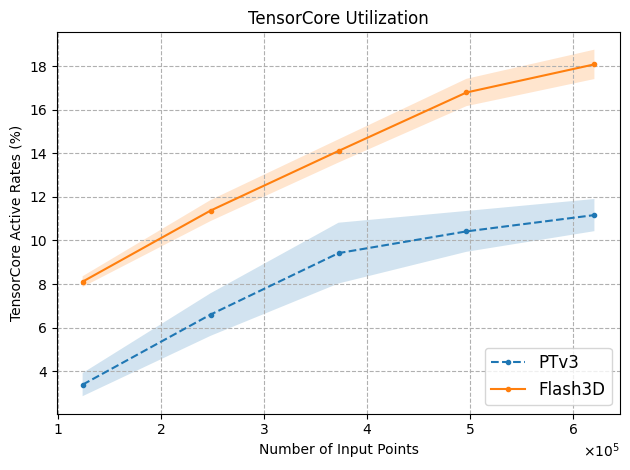}
    
    \vspace{-0.12in}
    \caption{TensorCore Active Rates vs. Input Sizes for Flash3D and PTv3.}
    \label{fig:tensorcore_active_rates}
    
    \vspace{-0.2in}
\end{figure}
\noindent \textbf{TensorCore Matrix Multiplication Utilization}
TensorCores contribute an overwhelming computing throughput to modern GPU chips 
\footnote{For H100 GPUs, FP16 TensorCores account for 1979 teraFLOPS among total peak computing capacities while traditional FP32 CUDA Cores account for 67 teraFLOPS.}.
TensorCore saturation improvements have amplified impacts on overall throughputs.
PTv3 struggles with less than 5\% TensorCore utilization in most point cloud scenarios, \textbf{wasting over 95\%} of GPU resources and investments. 
In contrast, Figure~\ref{fig:tensorcore_active_rates} highlights Flash3D overcomes these bottlenecks.
\begin{figure}[h]
    \vspace{-0.08in}
    \centering
    \includegraphics[width=\linewidth,height=2.2in]{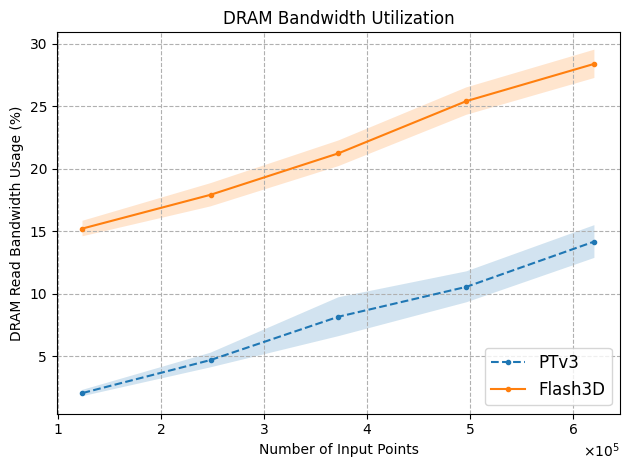}
    
    \vspace{-0.18in}
    \caption{DRAM Read Bandwidth Usage vs. Input Sizes for Flash3D and PTv3.}
    \label{fig:dram_bandwidth_usage}
    
    \vspace{-0.18in}
\end{figure}

\noindent \textbf{Memory Bandwidth Utilization (DRAM Read)}
Finally, we evaluate memory read bandwidth utilization (Figure~\ref{fig:dram_bandwidth_usage}). 
Flash3D excels in memory bandwidth usage metric by respecting \textit{memory locality}.
PTv3 bottlenecks on memory bandwidth due to its multiple global scattering operations.

\vspace{-0.1in}
\section{Conclusion}
By bridging the gap between algorithm development and hardware optimization, Flash3D Transformer opens new avenues for point cloud backbones with fast training and inference. 
Flash3D leverages insights from the joint hardware-geometry locality and outperforms previous state-of-the-art point transformer while achieving 2.25x speedup and 2.4x memory efficiency.
Our work underscores the importance of co-designing algorithms with hardware architectures to fully exploit the capabilities of modern GPUs.

\newpage

\clearpage
\setcounter{page}{1}
\maketitlesupplementary

\section{Outdoor 3D Tasks}
In this section, we provide more benchmark results on semantic segmentation tasks, including nuScenes~\cite{caesar2020nuscenes,fong2022panoptic} and Waymo open dataset~\cite{sun2020scalability}.
We also provide ablation results by varying the attention scope sizes.

\subsection{nuScenes Semantic Segmentation}
We present the benchmarks on nuScenes semantic segmentation for both validation set and test set.
Similar to Section~\ref{sec:exp}, we train two variants of Flash3D by fixing it under the same number of parameters and the same total memory costs.
We present the results in Table~\ref{tab:nuscenes_semseg_val_test}.
Our results show that Flash3D outperforms previous state-of-the-art results on both validation set and test set.
When we increase the memory budgets for Flash3D to include more parameters, Flash3D further improves the mIoU performance.
\begin{table}[ht]
\centering
\small
\begin{tabular}{lcc}
\toprule[1.2pt]
Methods & nuScenes Val & nuScenes Test \\
\midrule
MinkUNet~\cite{choy20194d}        & 73.3  & -  \\
SPVNAS~\cite{tang2020searching}          & 77.4  & -  \\
Cylinder3D~\cite{zhou2020cylinder3d}      & 76.1  & 77.2 \\
AF2S3Net~\cite{cheng20212}        & 62.2  & 78.0 \\
SphereFormer~\cite{lai2023spherical}    & 78.4  & 81.9 \\
PTv2~\cite{wu2022point} & 80.2  & 82.6 \\
PTv3~\cite{wu2024point} & 80.4  & 82.7 \\
Flash3D (Same param.)   & 81.2  & 83.1 \\
Flash3D (Same memory)   & 81.5  & 83.6 \\

\bottomrule[1.2pt]
\end{tabular}
\caption{Outdoor semantic segmentation on nuScenes validation and test sets.}
\label{tab:nuscenes_semseg_val_test}
\end{table}

\subsection{Waymo Semantic Segmentation}
\paragraph{Waymo Validation Set}
We benchmark two variants of Flash3D on Waymo Open Dataset semantic segmentation task in Table~\ref{tab:waymo_val}.
Flash3D consistently outperforms previous state-of-the-art results on both mIoU and mAcc metrics.
Flash3D demonstrates further scaling when we add more parameters.

\begin{table}[h]
    \centering
    \small
    \begin{tabular}{lcc}
        \toprule
        Methods & mIoU & mAcc \\
        \midrule
        MinkUNet~\cite{choy20194d} & 65.9 & 76.6 \\
        SphereFormer~\cite{lai2023spherical} & 69.9 & - \\
        PTv2~\cite{wu2022point} & 70.6 & 80.2 \\
        PTv3~\cite{wu2024point} & 71.3 & 80.5 \\
        Flash3D (Same param.) & 71.7 & 80.9 \\
        Flash3D (Same memory) & 72.5 & 81.6 \\
        \bottomrule
    \end{tabular}
    \caption{Waymo Val mIoU and mAcc comparison.}
    \label{tab:waymo_val}
\end{table}

\paragraph{Scaling up Attention Scopes}
On nuScenes semantic segmentation task, we scale attention scope sizes at 1024, 4096, and 8192 to benchmark PTv3~\cite{wu2024point} and two variants of Flash3D in Table~\ref{tab:attention_scope_nuscenes}.

PTv3 performance \textit{degrades} when we scale up the attention scope sizes.
When we fix the parameter size, Flash3D performance peaks at 4096.
And increasing attention scopes to 8192 degrades the task performance.

When we fix the memory budgets and add more parameters to Flash3D, increasing attention scope sizes improves the task performance.
\begin{table}[h]
    \centering
    \small
    \begin{tabular}{lccc}
        \toprule
        Attention Scope & 1024 & 4096 & 8192 \\
        \midrule
        PTv3~\cite{wu2024point} & 80.4 & 80.2 & 79.1\\
        Flash3D (Same param.) & 80.6 & 81.2 & 80.1\\
        Flash3D (Same memory) & 80.2 & 81.5 & 81.7\\
        \bottomrule
    \end{tabular}
    \caption{Attention Scope impacts on nuScenes semantic segmentation on the validation set }
    \label{tab:attention_scope_nuscenes}
\end{table}

We observe a similar trend when scaling up the attention scopes on Waymo validation set.
The improvements of scaling up attention scopes of Flash3D are more pronounced on Waymo validation set since Waymo has denser point clouds and more data.

\begin{table}[h]
    \centering
    \small
    \begin{tabular}{lccc}
        \toprule
        Attention Scope & 1024 & 4096 & 8192 \\
        \midrule
        PTv3~\cite{wu2024point} &   70.8 & 70.5 & 70.2\\
        Flash3D (Same param.)   &   71.5 & 71.7 & 71.3\\
        Flash3D (Same memory)   &   71.6 & 72.1 & 72.5\\
        \bottomrule
    \end{tabular}
    \caption{Attention Scope impacts on Waymo semantic segmentation on the validation set mIoU.}
    \label{tab:patch_size}
\end{table}

\section{Hardware Scalability}
\subsection{H100 Training and Inference Costs}
We present training and inference costs of Flash3D.
Similar to Section~\ref{sec:exp}, we train two variants of Flash3D: one with the same parameter sizes as PTv3~\cite{wu2024point}, and one with the same inference memory costs as PTv3~\cite{wu2024point}.
We present training latencies and inference latencies on H100 in Table~\ref{table:h100_comparison_resource}.
For fair comparisons, we report training latencies as the time to train an iteration when batch size is 1.
\begin{table}[h!]
    \centering
    \small 
    \setlength{\tabcolsep}{6pt} 
    \arrayrulecolor[gray]{0.2} 
    \begin{adjustbox}{max width=\columnwidth} 
        \begin{tabular}{>{\columncolor[gray]{0.92}}l c c c c c}
            \specialrule{1.5pt}{0pt}{0pt} 
            Scalability & \multicolumn{5}{c}{H100} \\
            (nuScenes) & Params. & Memory & Inf Latency & Train Latency & mIoU  \\
            \hline
            PTv3~\cite{wu2024point}    & \cellcolor[gray]{0.8} 46.2M & 1.2G & 30.2ms  &  77.6ms       & 80.4         \\
            Flash3D & \cellcolor[gray]{0.8} 46.2M & 0.5G & \textbf{13.4ms} & \textbf{33.8ms} & \textbf{81.2}\\
            \hline
            PTv3~\cite{wu2024point}    & 46.2M     & \cellcolor[gray]{0.8} 1.2G & 30.2ms &  77.6ms  & 80.4          \\
            Flash3D & 129.4M    & \cellcolor[gray]{0.8} 1.2G & \textbf{15.1ms} & \textbf{37.9ms} & \textbf{81.5} \\
            \specialrule{1.5pt}{0pt}{0pt} 
        \end{tabular}
    \end{adjustbox}
    \caption{Model scalability comparisons of PTv3 and Flash3D on H100 GPUs by fixing model parameter sizes and memory quotas respectively. Dark cells indicate fixed budgets. We fix attention scopes of all Flash3D models at 4096. mIoU indicates the semantic segmentation performance on nuScenes validation set.}
    \label{table:h100_comparison_resource}

\end{table}
\subsection{H100 Utilization Profiling}
In Section~\ref{sec:exp}, we describe key metrics and evaluate hardware scalability of Flash3D on \textbf{General Compute}, \textbf{TensorCore Matrix Multiplication}, and \textbf{Memory Bandwidth}.
In this section, we further elaborate on the implications of the metrics and present profiling results on H100 GPUs.
We keep the same setting as Section~\ref{sec:exp} to fix PTv3 and Flash3D at the same parameter sizes for profiling results.
Generally speaking, H100s are more memory bandwidth-starved and Flash3D shows further improvements over PTv3~\cite{wu2024point} on key metrics due to proper treatment of \textit{memory locality}.

\paragraph{General Compute}

\begin{figure}[h]
    \centering

    \includegraphics[width=\linewidth
    ]{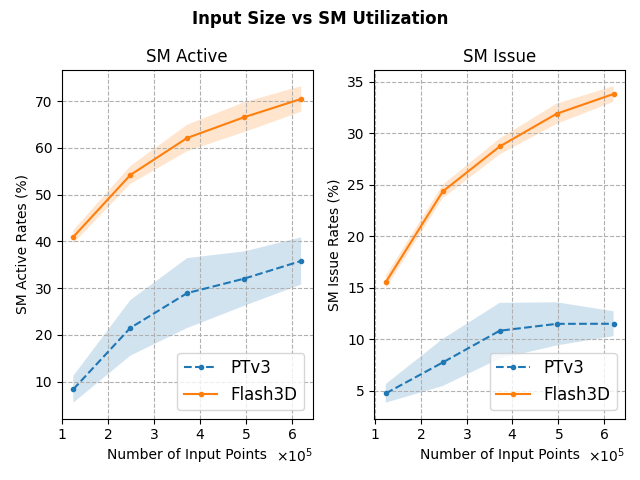}

    \caption{SM Utilization vs. Input Sizes for Flash3D and PTv3. We show the overall SM active rates on the left and more specific SM issuing rates on the right.}
    \label{fig:sm_active_rates_h100}
    
\end{figure}
SM cores are the \textit{computing tiles} for GPUs and H100 has 132 SM cores\footnote{Different versions of H100 offer options of 144 SMs, 132 SMs, and 113 SMs.}.
The aggregated SM utilization is a key measure of overall usage.
For more detailed illustrations of GPU architectures and hierarchies, refer to FlashAttention~\cite{dao2022flashattention} and \textsc{ThunderKittens}~\cite{spector2024thunderkittens}.

SM issuing rates focus more on the instruction front-end utilization.
A low SM issuing rate indicates the workloads are not properly divided among threads or the computing units are waiting for data.
Flash3D carefully spreads workloads among threads and SMs and demonstrates significant improvements on SM issuing rates in Figure~\ref{fig:sm_active_rates_h100}.

\begin{figure}[h]
    \centering
    \includegraphics[width=\linewidth]{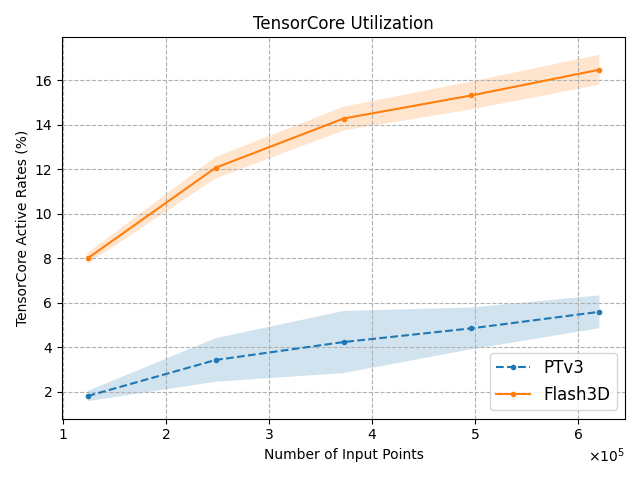}
    
    \caption{TensorCore Active Rates vs. Input Sizes for Flash3D and PTv3.}
    \label{fig:tensorcore_active_rates_h100}
    
\end{figure}
\paragraph{TensorCore Matrix Multiplication}
H100 TensorCores demand more data from other components of the GPU to saturate.
H100 poses harder barriers than A100 does in terms of TensorCore saturation~\cite{spector2024thunderkittens}.
PTv3 consistently demonstrates TensorCore utilization less than 5\%, which \textbf{wastes over 95\%} of GPU resources and investments as shown in Figure~\ref{fig:tensorcore_active_rates_h100}.

\begin{figure}[h]
    \centering
    \includegraphics[width=\linewidth]{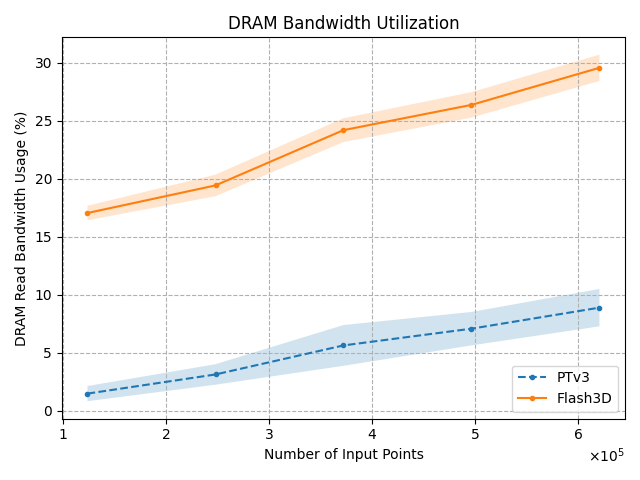}
    
    \caption{DRAM Read Bandwidth Usage vs. Input Sizes for Flash3D and PTv3.}
    \label{fig:dram_bandwidth_usage_h100}
    
\end{figure}

\paragraph{Memory Bandwidth}
H100 computing throughput disproportionally improves over its memory bandwidth~\cite{spector2024thunderkittens}.
Therefore, H100 demands more realized memory bandwidth to keep its computing tiles working.
Flash3D carefully treats memory locality and enjoys a scalable usage of memory bandwidth, shown in Figure~\ref{fig:dram_bandwidth_usage_h100}.

\section{Implementation Details}
In this section, we further elaborate on implementation details and algorithm designs from Section~\ref{sec:method} and Section~\ref{sec:exp}.
We will open source our implementations after the anonymous review period.

\subsection{PSH with Two-Stage Counters}
In our main PSH algorithms, we used \texttt{AtomicInc}, \texttt{AtomicDec} on globally visible memory (DRAM), which establishes a global linear order among all threads on a GPU, and hence contiguous and unique indices for all points.
We implemented our naive version of PSH algorithms as described in Section~\ref{sec:method}.
The naive version demonstrated acceptable latencies, roughly at 0.4ms for 120k points.

However, our rebalancing algorithm generates a significant amount of memory traffic requiring global atomic guarantees, which throttles the overall throughput.
Therefore, we instead implemented a {\em two-stage-counter} version of our PSH algorithms, where we temporarily break the global linear order and coalesce temporary copies in batches.

Specifically, we create temporary copies of \textbf{bkt\_ctr} within local L1 cache blocks (shared memory), which allow fast localized atomics, which we will refer to as \texttt{AtomicInc\_block} and \texttt{AtomicDec\_block}.
We usually have 256 buckets and the counters typically have a shape of $\texttt{UInt32}[256]$, costing 1024 bytes, so they fit in local L1 cache blocks.

We break our original counter \texttt{AtomicInc} into three steps: local \texttt{AtomicInc\_block}, bulk commit to global memory \texttt{AtomicAdd}, and rebasing \textbf{bucket\_offset}:

\paragraph{Local Atomic Operations}
We initialize local copies \textbf{bkt\_ctr\_local} with zeros. 
When a GPU thread representing a point needs \texttt{AtomicInc}, it should $\texttt{AtomicInc\_block}(\textbf{bkt\_ctr\_local})$ to obtain \textbf{bucket\_off\_local}.

\paragraph{Bulk Commit}
We hold all threads in a block to a block-wise synchronization point and make sure every thread has finished their \texttt{AtomicInc\_block}.
Then we $\texttt{AtomicAdd}(\textbf{bkt\_ctr}, \textbf{bkt\_ctr\_local})$ to obtain \textbf{bkt\_rebase\_offset}.
Recall that \textbf{bkt\_ctr\_local} starts from zeros, so it represents a contiguous range of indices based on a future global version to be determined.
When we \texttt{AtomicAdd}(\textbf{bkt\_ctr}, \textbf{bkt\_ctr\_local}), we establish the starting point of the specific local range of indices in terms of the global counters.
At the same moment, we obtain a snapshot of global counters before \texttt{AtomicAdd}, which are exactly the bases for this \textbf{bkt\_ctr\_local}.
Therefore, we refer to this global counter snapshot as \textbf{bkt\_rebase\_offset}.

\paragraph{Rebasing \textbf{bucket\_offset}}
Recall that we have \textbf{bucket\_off\_local} in the first step, which needs to be adjusted and merged to the global linear order.
Similar to our discussion in the second step, \textbf{bucket\_off\_local} represents an offset starting from an undetermined global point. 
At the end of the second step, we determine this global point as \textbf{bkt\_rebase\_offset}.
Therefore, we can finalize $\textbf{bucket\_offset} = \textbf{bucket\_off\_local} + \textbf{bkt\_rebase\_offset}$.

In Section~\ref{sec:exp}, we report metrics from our {\em two-stage-counter} PSH algorithms.
Our {\em two-stage-counter} PSH algorithms further localize memory traffics within \textit{GPU tiles} and boost the throughput.

\subsection{ThunderKittens}
\textsc{ThunderKittens}~\cite{spector2024thunderkittens} is a CUDA library that provides convenient tools to implement DL kernels that demand TensorCore throughputs, such as FlashAttention algorithms~\cite{dao2022flashattention,dao2023flashattention,shah2024flashattention}.
Original FlashAttention algorithms have highly complex implementations and require lots of engineering hours for proper implementations\footnote{\url{https://github.com/Dao-AILab/flash-attention}}.

\textsc{ThunderKittens}~\cite{spector2024thunderkittens} simplifies FlashAttention implementations by providing numerical operations and matrix multiplications on $16\times 16$ matrix tiles. Our implementation uses an early version of \textsc{ThunderKittens} released on May 2024\footnote{\url{https://github.com/HazyResearch/ThunderKittens}}.

Current version of \textsc{ThunderKittens}~\cite{spector2024thunderkittens} incorporates a Load-Compute-Store-Finish (LCSF) pipeline to further optimize for newer GPUs including H100.
LCSF rectifies the asynchronous producer-consumer paradigm of \textsc{ThunderKittens}.
With LCSF updates, \textsc{ThunderKittens} can oversubscribe resources within SM, better saturate TensorCores, and outperform FlashAttention-3 in several settings~\cite{shah2024flashattention,spector2024thunderkittens}.

Unfortunately, Flash3D builds on top of FlashAttention-2~\cite{dao2023flashattention} and an early version of \textsc{ThunderKittens}, and we did not have a chance to incorporate the latest version of \textsc{ThunderKittens}.
We aim to incorporate newer \textsc{ThunderKittens} and FlashAttention-3~\cite{shah2024flashattention} in our future work to further boost Flash3D throughput and efficiency.

\subsection{Bucket-Swin}
We describe the implementation details of our \textbf{Bucket-Swin} attention based on \textsc{ThunderKittens}.
We focus on explaining the zero-overhead nature of \textbf{Bucket-Swin} attention.
By ``zero-overhead," we mean that our \textbf{Bucket-Swin} attention only incurs MHSA costs without additional memory or latency. Since \textsc{ThunderKittens} loads inputs in tile-sized chunks, we achieve zero-overhead bucket-swin by simply loading into L1 the tiles corresponding to the set of buckets among which we wish to compute attentions, as detailed in the following paragraphs.

After our PSH algorithms, point features are represented in a contiguous array $\texttt{FP16}[N, d]$, where $N$ is the number of points and $d$ is the number of feature dimensions.
Our point feature array is a concatenation of buckets along the $N$ dimension.
Therefore, any bucket-size-aligned sub-array represents a bucket.
For example, when bucket size is 512, a sub-array $\texttt{FP16}[512:1024, d]$ contains all feature vectors of a bucket.

The goal of \textbf{Bucket-Swin} attention is to associate arbitrary buckets to an attention scope and compute MHSA within this logical scope.
Notably, our \textbf{Bucket-Swin} attention does \textbf{not} introduce additional shuffling of the feature array.
For a point $i$, its input features are located at $\texttt{FP16}[i, d]$ and its output features are located at $\texttt{FP16}[i, d]$ as well.
Such fixed layout is the key to our zero-overhead \textbf{Bucket-Swin} attention.\footnote{In Figure~\ref{fig:flash3d_dataflow_overview}, we illustrate bucket-swin by pointing memory blocks to the destination memory blocks to best convey the conceptual model. In practice, we don't incur physical memory block rewriting but instead implement address redirection during FlashAttention-2 computation. Therefore, we have fixed memory layout and zero-overhead bucket-swin attention.}

We describe how to vary attention scopes without shuffling point feature indices.
As described above, a bucket-aligned sub-array represents a bucket.
Consider an example of computing MHSA among two buckets: $B_i$ and $B_j$, where $B_i, B_j$ include contiguous point indices of buckets $i,j$ respectively.
Before MHSA, we map point features $\texttt{FP16}[i, d]$ into $Q\in\mathbb{R}^{N\times d}$, $K\in\mathbb{R}^{N\times d}$, and $V\in\mathbb{R}^{N\times d}$.
Point features of $B_i$ are $\texttt{FP16}[B_i, d]$.
Query-key-value triplets of $B_i$ are $Q[B_i, d]$, $K[B_i, d]$, and $V[B_i, d]$.
We enforce that bucket sizes are all multiples of 16 so $Q[B_i, d]$, $K[B_i, d]$, and $V[B_i, d]$ can be tiled into $16\times 16$ sub-matrices by \textsc{ThunderKittens}.

FlashAttention-2~\cite{dao2023flashattention} operates on $16\times 16$ tiles.
As long as memory layouts for $Q, K, V$ arrays suit $16\times 16$ tiling, FlashAttention-2 operates on the original principles and assumptions.
In our example, $Q, K, V$ arrays for buckets $B_i$ and $B_j$ support $16\times 16$ tiling.
Therefore, we have full knowledge of tile addresses for buckets $B_i$ and $B_j$.
To compute MHSA output for $B_i$ under a scope of $\{B_i, B_j\}$, we fetch tiles from $Q[B_i, d]$, $K[B_i, d]$, $K[B_j, d]$, $V[B_i, d]$, $V[B_j, d]$ to L1 cache blocks.
The fetching step does not incur overheads other than plain FlashAttention-2~\cite{dao2023flashattention}.
The only difference to FlashAttention-2 is redirecting tile addresses according to our bucket scheme.
Then the computation and output steps are the same as FlashAttention-2.
Therefore, our \textbf{Bucket-Swin} attention incurs no overheads on top of FlashAttention-2.

\subsection{In-bucket Pooling}
In Section~\ref{sec:method_tsfmr_stage}, we motivate to replace the grid pooling operation~\cite{wu2022point} by our in-bucket pooling to capitalize on the existing geometric locality of buckets.
Our in-bucket pooling provides higher throughput by removing global serialization and neighborhood finding.
In addition, our in-bucket pooling has a fixed pooling ratio to offer smoother memory access patterns that can be achieved on GPU tiles.
In this section, we describe further algorithm and implementation details of our in-bucket pooling.

Our main PSH algorithm establishes a linear order based on globally visible counters.
However, our in-bucket pooling sub-bucket construction operates within local SMs (1024 points for a ThreadBlock).
In contrast to our main PSH rebalancing algorithm, our in-bucket pooling sub-bucket construction includes three steps: \textbf{initial sub-buckets}, \textbf{new sub-bucket allocation}, \textbf{find new sub-bucket}.

\paragraph{Initial sub-buckets}
Similar to our main PSH algorithm, we use a hash function to assign each point a \textbf{subbuck\_id}.
We keep a ThreadBlock counter to allocate \textbf{subbuck\_id}.
Since points have uneven distributions among sub-buckets, sub-buckets have points no less than the desirable capacity $\rho$.
Similarly, allocated \textbf{subbuck\_id} is no more than the desirable total sub-buckets.

\paragraph{New sub-bucket allocation}
Then we examine each sub-bucket to relocate points beyond the requested capacity $\rho$.
For each point that's beyond the sub-bucket capacity $\rho$, we allocate a new sub-bucket with id \textbf{subbuck\_id}.
We treat all points until \textbf{subbuck\_id} reaches the total sub-bucket number.
At this point, sub-buckets allocated in the first step should contain exactly $\rho$ points.
And each newly allocated sub-bucket should contain 1 point.

\paragraph{Find new sub-bucket}
In this step, we treat the remaining points in this ThreadBlock.
For each point, we scan all newly allocated sub-buckets and find one that has points closest to its coordinate.
Then we try to commit this point into the newly found sub-bucket by \texttt{AtomicInc}.
However, a sub-bucket might have reached its capacity before this point joins.
It manifests in the result returned by \texttt{AtomicInc} being greater than or equal to $\rho$.
Then we synchronize all threads to identify filled sub-buckets and under-filled sub-buckets.
Finally, we repeat this step until all points find a sub-bucket.

Note that we repeat for multiple times of linear scans of newly allocated sub-bucket, which might seem to be a significant cost.
However, all of our in-bucket pooling algorithms are confined within SMs running on fast local L1 cache blocks.
Operations on local L1 cache blocks are significantly faster than those interacting with globally visible memory.
Hence, our sub-bucket construction has higher throughput than our main PSH algorithm.

Based on our sub-bucket construction, we reduce point features within each sub-bucket to complete our in-bucket pooling operations as describe in main sections.

\newpage

{
    \small
    \bibliographystyle{ieeenat_fullname}
    \bibliography{main_arxiv}
}


\end{document}